\documentclass[a4paper, 10pt]{LTJournalArticle}
\usepackage{amsmath}
\usepackage{graphicx}
\usepackage{subcaption}
\usepackage{stfloats}
\title{End-to-End Lidar-Driven \\ Reinforcement Learning for Autonomous Racing}
\author{Meraj Mammadov\thanks{Corresponding author: \href{mailto:meraccos@gmail.com}{meraccos@gmail.com}}}

\date{\footnotesize Department of Mechanical Engineering, Ulsan National Institute of Science and Technology}

\renewcommand{\maketitlehookd}{%
	\begin{abstract}
		\noindent Reinforcement Learning (RL) has emerged as a transformative approach in the domains of automation and robotics, offering powerful solutions to complex problems that conventional methods struggle to address. In scenarios where the problem definitions are elusive and challenging to quantify, learning-based solutions such as RL become particularly valuable. One instance of such complexity can be found in the realm of car racing, a dynamic and unpredictable environment that demands sophisticated decision-making algorithms. This study focuses on developing and training an RL agent to navigate a racing environment solely using feedforward raw lidar and velocity data in a simulated context. The agent's performance, trained in the simulation environment, is then experimentally evaluated in a real-world racing scenario. This exploration underlines the feasibility and potential benefits of RL algorithm enhancing autonomous racing performance, especially in the environments where prior map information is not available.
	\end{abstract}
}

\begin{document}

\maketitle 


\section{Introduction}

The quest for autonomous driving has intrigued researchers across various disciplines, including artificial intelligence, control systems, and robotics\autocite{Huang:2020}. Central to these studies is the challenge of navigating complex and dynamic environments. This complexity becomes particularly pronounced in autonomous racing, where traditional solutions have extensively relied on methods such as precise localization and mapping, thorough path planning, and rule-based decision-making systems.

Conventional localization and mapping techniques, such as Simultaneous Localization and Mapping (SLAM), typically involve intensive computations, high-quality sensory data, and substantial hardware requirements. Similarly, rule-based systems and path planning algorithms require exhaustive definitions of behaviors and scenarios. These methodologies, although proven in scenarios with known and structured environments, can struggle significantly in situations where there is limited or no prior information about the complexity of the map\autocite{Merzlyakov:2021}. This inherent inflexibility underscores the necessity for methods that can generalize to a wide spectrum of scenario and environments. 

In recent years, Reinforcement Learning (RL) has emerged as a compelling solution to such dynamic problems\autocite{Piotr:2017, Julio:2020, Maximilian:2018}. The strength of RL lies in its ability to generate solutions to a wide range of scenarios without the need for tailoring specific solutions for each case. It achieves this by learning from extensive interaction with the environment, optimizing its policies based on the reward signals. The interaction with diverse examples and the approximation nature of the neural networks allow the RL agent to generalize its solutions to new, unseen scenarios.

In the context of car racing, an RL agent can be trained on immediate sensory data (e.g., camera output) to understand the environment and make proper driving decisions, thereby bypassing the need for global localization and the challenges of the conventional methods. 

In a relevant previous work, Maximilian et al.\autocite{Maximilian:2018} has shown that it is possible to train an end-to-end DRL agent using camera sensors to safely navigate through highways. One major shortcoming of using camera sensor for this regard lies in the lack of its generalization to different lighting and weather conditions, especially during nighttime. It is also nontrivial to match the synthetic camera outputs from the simulation to the real world, making it challenging to transfer the learned model to physical experimentation.

Considering these aspects, this work focuses on utilizing raw lidar and odometry readings to train an end-to-end RL agent to safely navigate through racing tracks. Once the training phase is completed in the simulation, the trained agent's ability to generalize to real-world scenarios is experimentally tested in small-scale racing tracks.

This paper is structured as follows: The first section delves into the reinforcement learning algorithm employed for training the autonomous racing agent. The second section provides an overview on the simulation environment used for the training process. The third section presents the domain randomization techniques applied during the training phase to enhance the agent's ability to adapt to a wide range of scenarios. The last section examines the agent's training results and behavior in both simulation and real-world experiments.

\section{Reinforcement Learning Algorithm}
\subsection{Problem Definition}
The task is to design a RL agent capable of navigating a race track while avoiding obstacles and other vehicles. The agent, at any given time step, is equipped with the following sensory inputs:
\begin{itemize}
    \item \textbf{Lidar data}: This serves as the agent's primary perception tool, allowing it to detect the boundaries of the environment and any obstacles within it.
    \item \textbf{Odometry data}: This provides the agent with knowledge of its current velocity, essential for executing safe and effective maneuvers.
    \item \textbf{Previous action}: The action commanded in the previous time step is given to the agent to offer context for understanding the evolving dynamics of the car.
\end{itemize}

These inputs are chosen to ensure that the problem retains the Markovian property, meaning the state at the next time step is conditionally independent of past states given the present state. This Markovian property allows the problem to be modeled as a Markov Decision Process (MDP)\autocite{Sutton:2023qr}.

An MDP is a tuple $(S, A, P, R)$ where:

\begin{itemize}
\item $S$ is the state space, the range of the observations the agent can experience.
\item $A$ is the action space, the range of the actions the agent can command.
\item $P: S \times A \times S \rightarrow [0, 1]$ is the transition function.
\item $R: S \times A \rightarrow \mathbb{R}$ is the reward function, which is designed to encourage the agent to learn expected behaviors.
\end{itemize}

Given this formalization, the RL algorithm's objective is to learn a policy $\pi: S \rightarrow A$, mapping states to actions, that maximizes the expected cumulative reward. The chosen actions should enable the agent to navigate safely through the racing environment while avoiding obstacles and other vehicles.

\subsection{Proximal Policy Optimization}
Proximal Policy Optimization (PPO) \autocite{Schulman:2017}{} is a widely used policy-based  on-policy reinforcement learning algorithm well-known for its high performance, robustness to hyperparameters and computational efficiency. PPO seeks to address the challenges of policy optimization in a simpler and effective manner compared to the previous methods like Trust Region Policy Optimization (TRPO) \autocite{Schulmann:2015}. 

The core concept of PPO is to limit the policy update step size to ensure stable and efficient learning. This is achieved through a specialized objective function, which discourages the policy from moving too far from the current policy. 

The algorithm introduces a surrogate objective function:
\begin{equation*}
    L(\theta) = E_t[\min(r_t(\theta)A_t, \text{clip}(r_t(\theta), 1-\epsilon, 1+\epsilon)A_t)]
\end{equation*}

Here, $\theta$ represents the policy parameters, $A_t$ is the advantage function at time step $t$, $r_t(\theta)$ is the likelihood ratio, and $clip$ is a function that limits the value of its first argument to be between $1-\epsilon$ $1+\epsilon$. The policy is updated by maximizing this objective function.

\subsection{Model Architecture}
Both the policy and value networks consists of only linear layers with two hidden layers. As a best practice, the value model was given a rather high number of neurons (64, 64). The policy network worked the best with smaller (32,32) model structure. Both models were given Tanh nonlinearity with orthogonal weight initialization \autocite{Hu:2020}.

\subsection{State and Action Space}
The state space is defined by three types of data: raw lidar readings (2155 data points), the magnitude of the velocity (single float), and the previous step's action (float array of speed and steering). The action space is composed of two float values, representing forward speed and steering angle (in radians).

To ensure more stable training and increased performance, it is a common practice to normalize inputs before feeding them into the neural network. For this regard, the lidar readings are clipped at 10m and normalized to a range between 0 and 1. The action values are also normalized to a range of -1 and 1.

To give the agent an understanding of the environment dynamics, the observation also includes the previous three step's observations, stacked together.

\subsection{Domain Randomization}
Domain randomization was employed to prevent model overfitting to specific characteristics of a single race track. The applied randomizations include:

\begin{itemize}
\item Training on approximately 500 distinct race tracks generated via polynomial fitting, with variations in track shape, curvature, length, and width.
\item Randomly placing obstacles of varying shapes and sizes along the track, replicating potential external objects and obstacles in the experimental setup.
\item Introducing random delay and noise into the lidar and velocity sensor readings, simulating real-world sensor imperfections and response latency.
\end{itemize}

\begin{figure*}[!tb]
  \centering
  \begin{subfigure}[b]{0.33\textwidth}
    \includegraphics[width=\textwidth, viewport=200 200 1400 1400, clip]{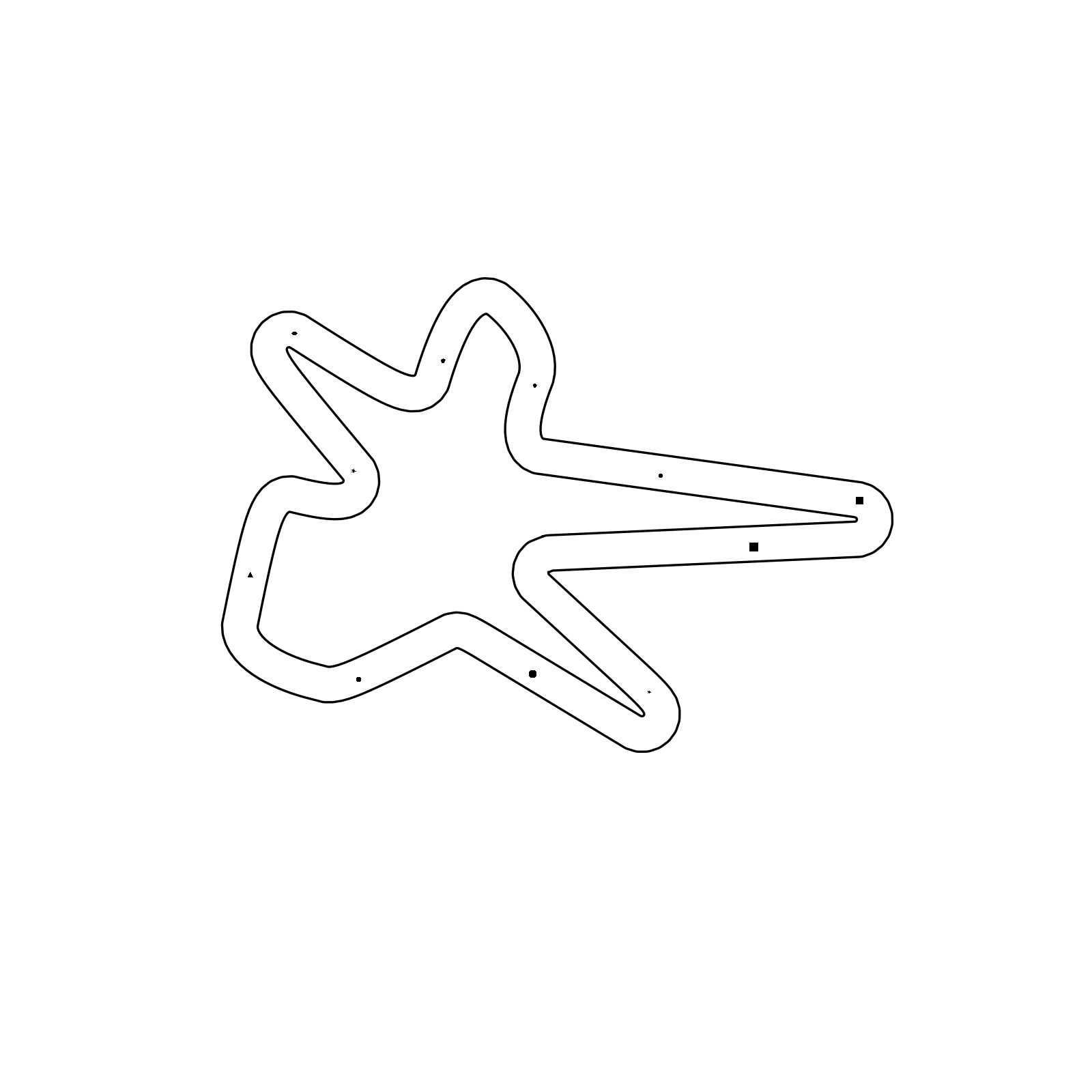}
  \end{subfigure}
  \hfill
  \begin{subfigure}[b]{0.33\textwidth}
    \includegraphics[width=\textwidth, viewport=200 200 1400 1400, clip]{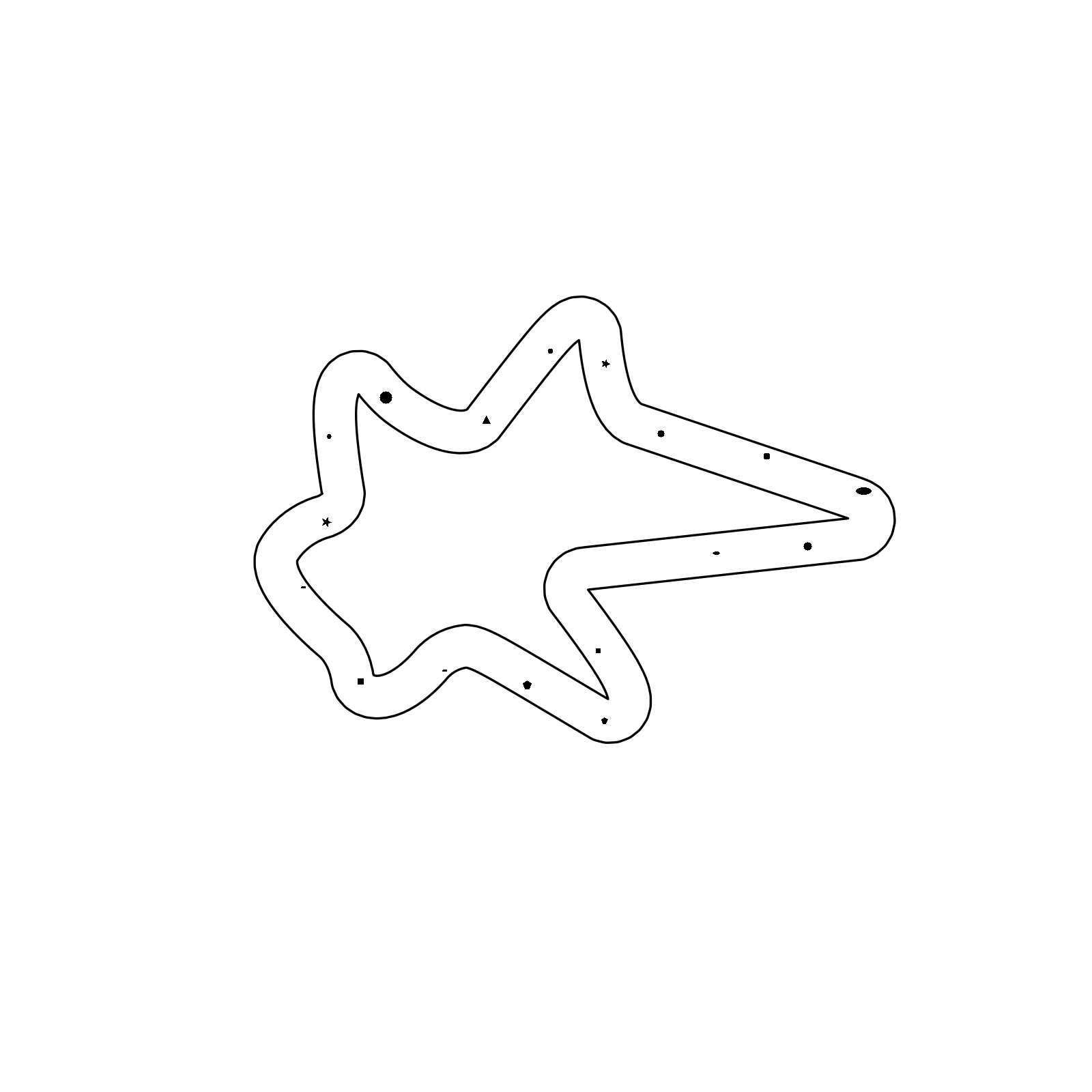}
  \end{subfigure}
  \hfill
  \begin{subfigure}[b]{0.32\textwidth}
    \includegraphics[width=\textwidth, viewport=200 200 1400 1400, clip]{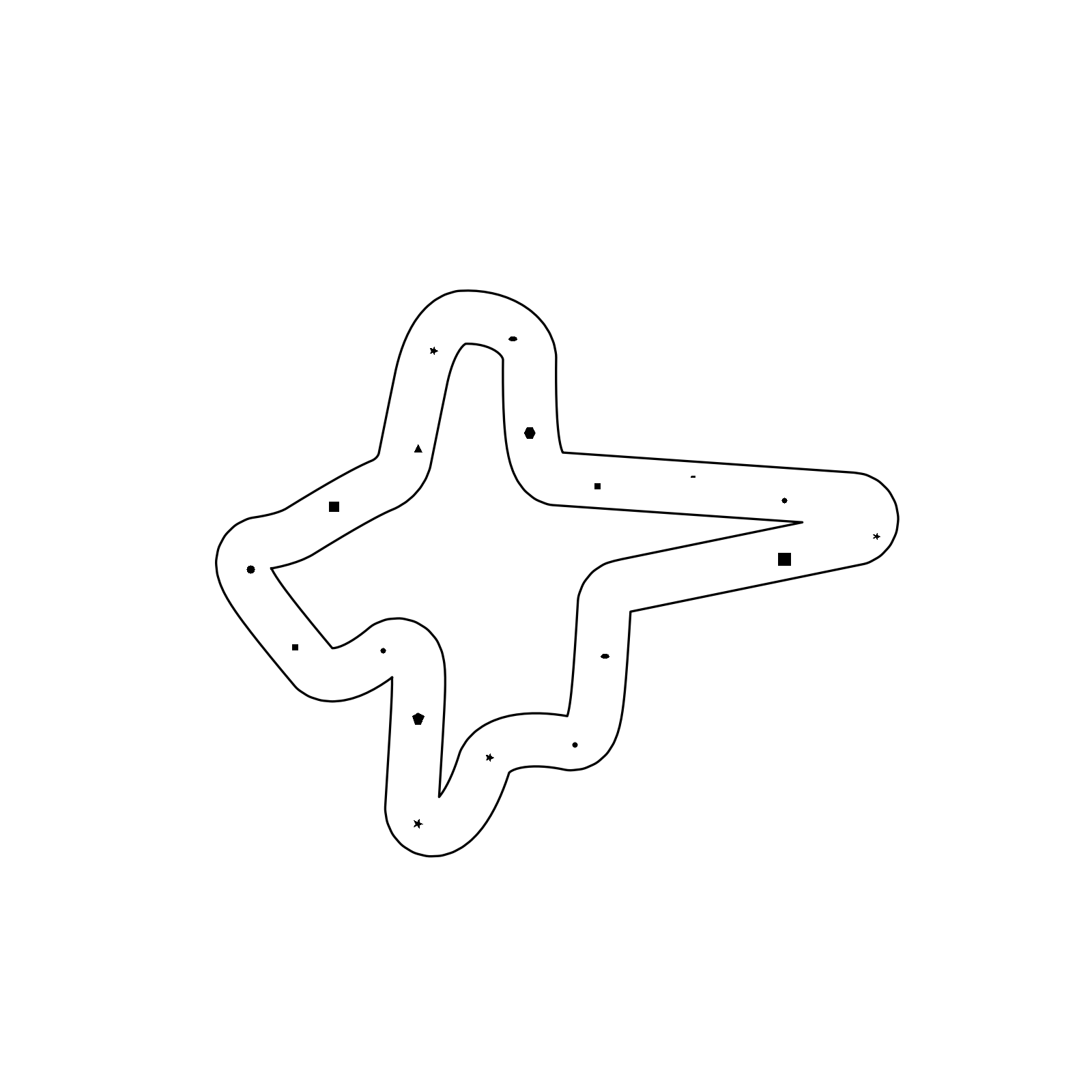}
  \end{subfigure}
  \caption{Three of the training maps with varying curvature and width. The random obstacles represent external real-world objects.}
  \label{fig:maps}
\end{figure*}

Figure \ref{fig:maps} present three of the maps used for the training, along with the randomly added obstacles.

\subsection{Reward Function}
The design of the reward function is a critical aspect of reinforcement learning. This function guides the agent's behavior by providing a quantitative representation of the desirability of different actions. For effective mapping of the race track, we translate the Cartesian (xy) coordinates into Frenet (sd) coordinates. The 's' represents the distance along the track centerline, and 'd' signifies the lateral displacement from the centerline. This transformation utilizes the centerline data of the race track maps.

The following are the key aspects of the agent's behavior that the reward function aims to promote or discourage:

\begin{itemize}
\item \textbf{$v_s$}: Reward increases with velocity along the track, encouraging the agent to maintain a high speed.
\item \textbf{$v_d$}: The reward decreases as the orthogonal velocity increases, discouraging lateral movement.
\item \textbf{d}: The closer the agent is to the centerline, the higher the reward, encouraging track centering.
\item \textbf{Steer action}: Higher steering actions yield lower rewards, promoting smooth, stable driving.
\item \textbf{Collision}: Collisions are highly penalized, discouraging contact with obstacles or boundaries.
\end{itemize}

Incorporating the above factors, the reward function under no collision is formulated as follows:
\begin{equation*}
    \text{reward} = 1.0v_s - 0.01\lvert v_d \rvert - 0.02\lvert d \rvert - 0.1\lvert action_{steer} \rvert
\end{equation*}

If a collision happens, the agent is penalized by a highly negative value: $\text{reward}=-1000$, and the episode is terminated.

\begin{figure}[!b] 
	\includegraphics[width=\linewidth]{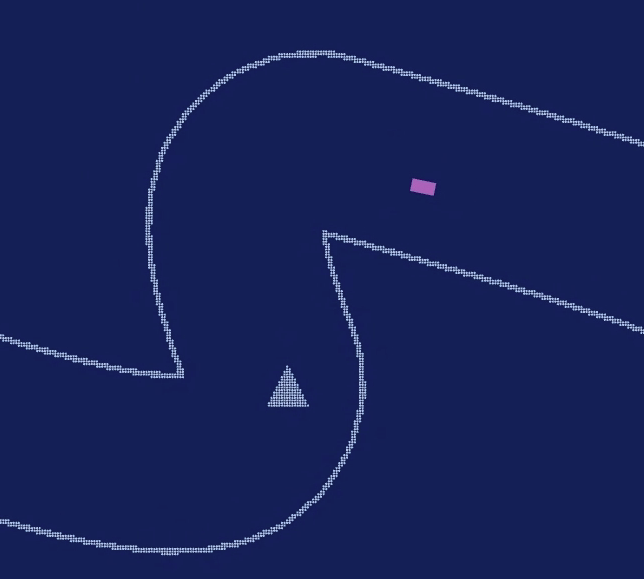}
	\caption{A snapshot from the simulation environment. The model car is driving through the race track while maintaining the centerline. The triangle is an example obstacle.}
	\label{fig:simulation}
\end{figure}

\section{Simulation Environment}

Due to the accessibility of the physical vehicle and the extant Gym environment, the open-source platform, F1TENTH simulation environment is employed in this experiment. To mimic the lidar sensor of the vehicle, it is developed in accordance with the actual hardware on the car. A major difference with the physical model is the dynamics: the simulator leverages a simplified bicycle model which, although similar, does not provide a perfect mirroring of the actual vehicular dynamics.

Figure \ref{fig:simulation} showcases an example scenario from the simulation environment. The tracks contain complex layouts and sharp turns that require nontrivial decisions to be made.

The full implementation of the simulation and training code can be found in \url{https://github.com/meraccos/f1tenth_reinforcement_learning}.

\begin{figure*}[!t]
  \centering
  \begin{subfigure}[b]{0.49\textwidth}
    \includegraphics[width=\textwidth]{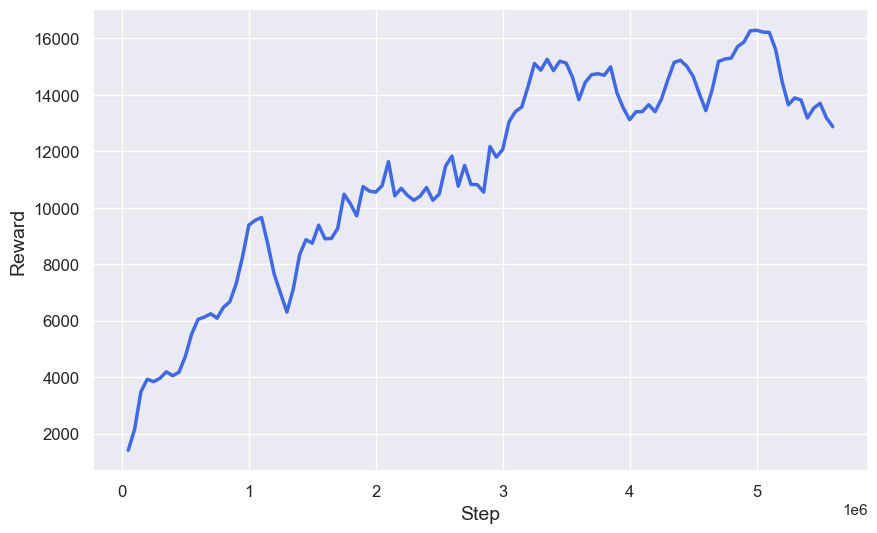}
    \caption{Reward trend}
  \end{subfigure}
  \hfill
  \begin{subfigure}[b]{0.49\textwidth}
    \includegraphics[width=\textwidth]{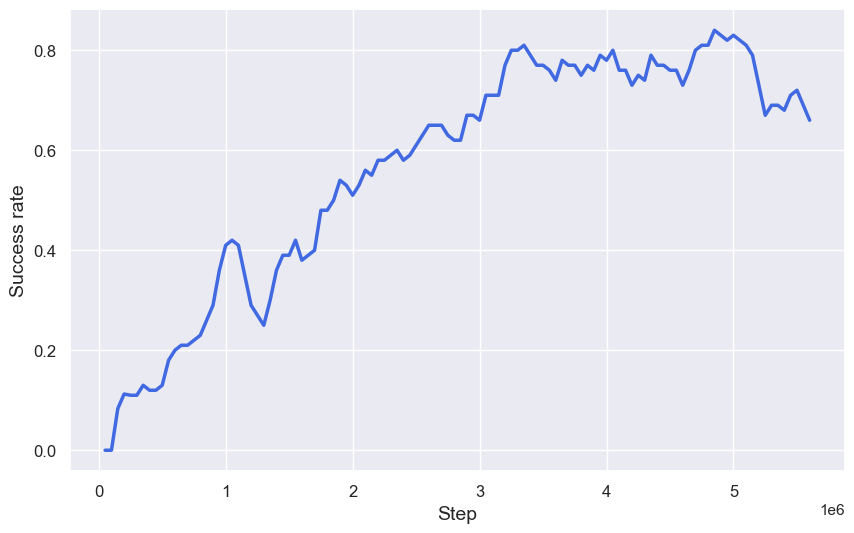}
    \caption{Success rate trend}
  \end{subfigure}
  \hfill
  \caption{The training results. (a) the episodic reward versus training time steps, steady increase. (b) the running mean of the successes over the episodes. Steady increase, similar to the reward trend. Stagnation at 80 percent.}
  \label{fig:training}
\end{figure*}


\section{Results}

\begin{table}[!b] 
	\caption{Hyperparameters used for the training}
	\centering
	\begin{tabular}{l r}
		\toprule
		\multicolumn{2}{c}{Hyperparameters} \\
		\cmidrule(r){1-2}
		Parameter  & Value \\
		\midrule
		Learning Rate  & 0.0001 \\
		Gradient clip  & 0.02 \\
        Entropy Coefficient & 0.0 \\
        Batch size  & 2048 \\
        Minibatch size & 256 \\
        Gamma & 0.998 \\
        Frame Stack & 4 \\
		\bottomrule
	\end{tabular}
	\label{tab:hyperparameters}
\end{table}

\subsection{Simulation Results}
This section presents the training results of the reinforcement learning agent trained for over 12-15 hours. Figure \ref{fig:training} (a) shows the reward curve of the training. The training process shows a progressive increase in the model performance.

To be able to compare different models more quantitatively, the success of the training agent is defined as its ability to complete a full lap without colliding with the track walls or randomly placed obstacles. The moving average of successful runs, calculated over the preceding 40 evaluation episodes, is visualized in Figure \ref{fig:training} (b). The success rate graph shows a close similarity with the reward trend, which once again proves that the designed reward properly motivates the agent to succeed the lap.

The trained agent demonstrates good performance in following the track and avoiding obstacles of different shapes. It has effectively learned to prioritize speed optimization while maintaining a centered position on the track. In scenarios presenting sharp turns or unclear sections, the agent adapts by reducing its speed and navigating cautiously until the path ahead is clear.

However, the agent's approach to obstacle avoidance occasionally appears to overly rely on the simulated dynamics, which do not perfectly resemble that of the real-world conditions. This discrepancy underscores the importance of refining the model's understanding of realistic dynamics for further improvement.

The training gets stagnated around 80 percent of success rate. The trained agent almost only fails in the cases with either extreme turns or when the randomly placed obstacles block the track.

The hyperparameters that resulted in the best performance are given in the Table \ref{tab:hyperparameters}.

\begin{figure*}[t]
  \centering
  \begin{subfigure}[b]{0.49\textwidth}
    \includegraphics[width=\textwidth]{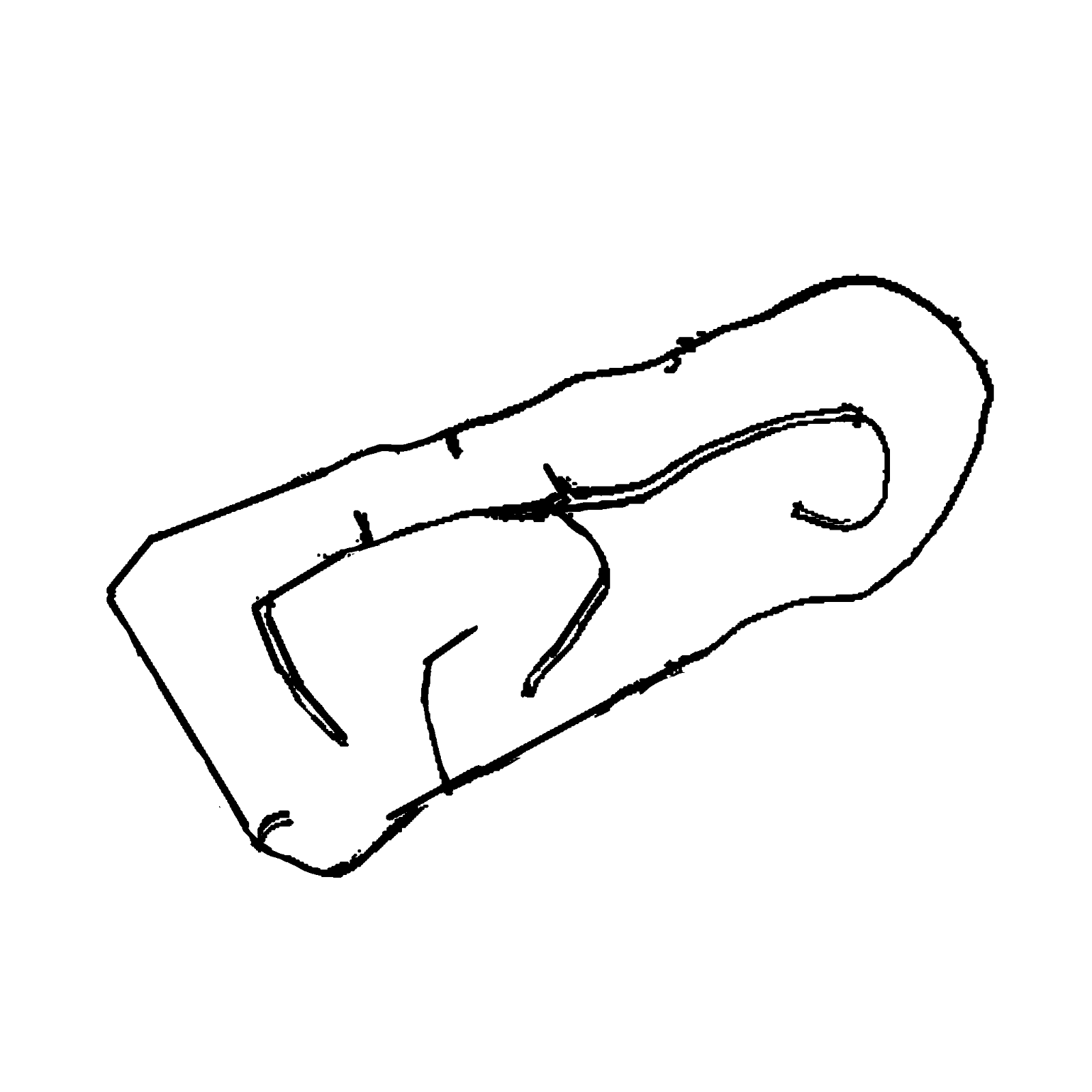}
    \caption{Birdview}
  \end{subfigure}
  \hfill
  \begin{subfigure}[b]{0.49\textwidth}
    \includegraphics[width=\textwidth]{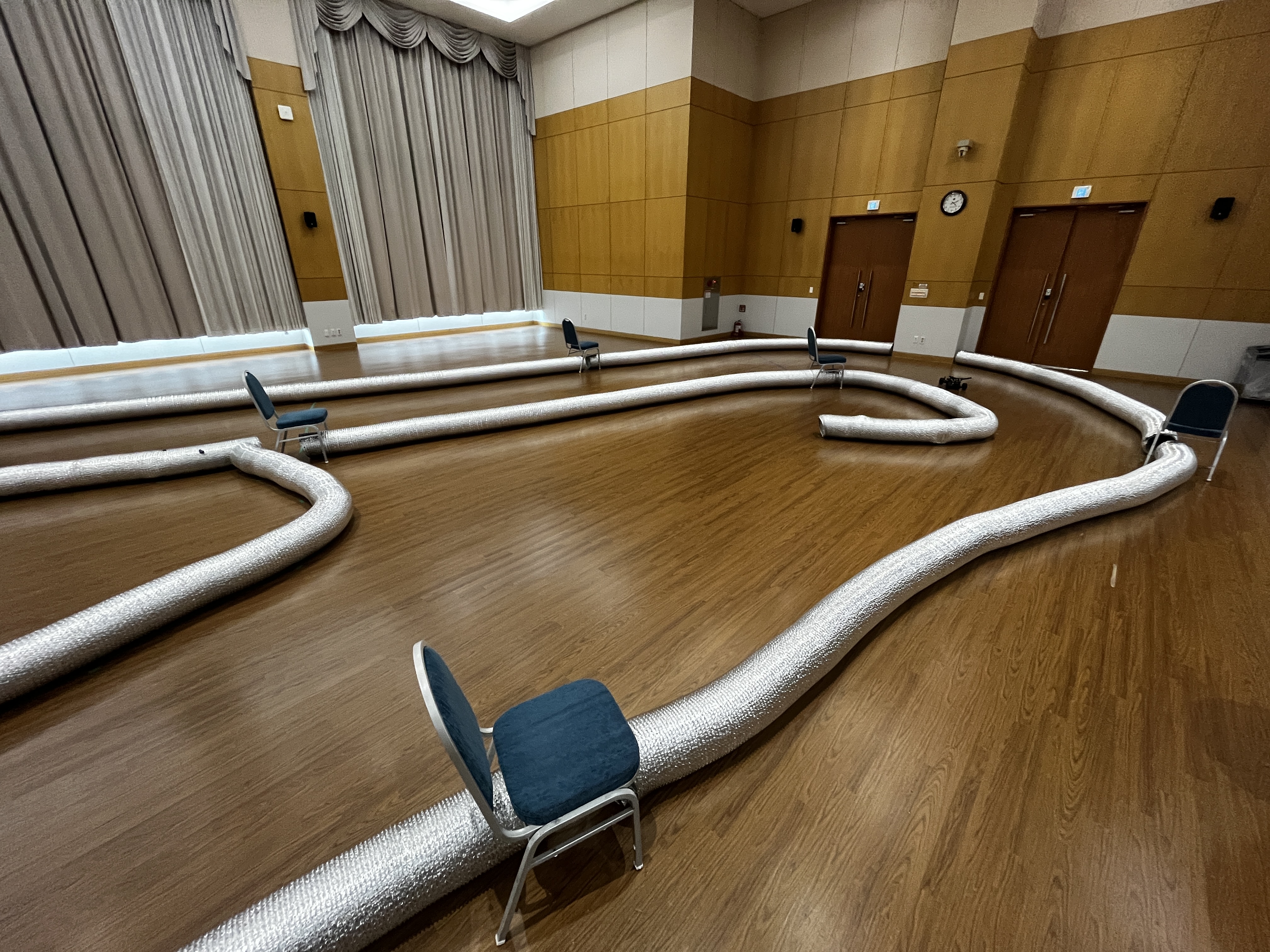}
    \caption{Scale}
  \end{subfigure}
  \caption{The map used for experimental testing. (a) shows the birdview borders of the map. (b) shows the scale of a portion of the map. The map contains sharp turns and challenging features.}
  \label{fig:realmap}
\end{figure*}

\begin{figure}[!b] 
	\includegraphics[width=\linewidth]{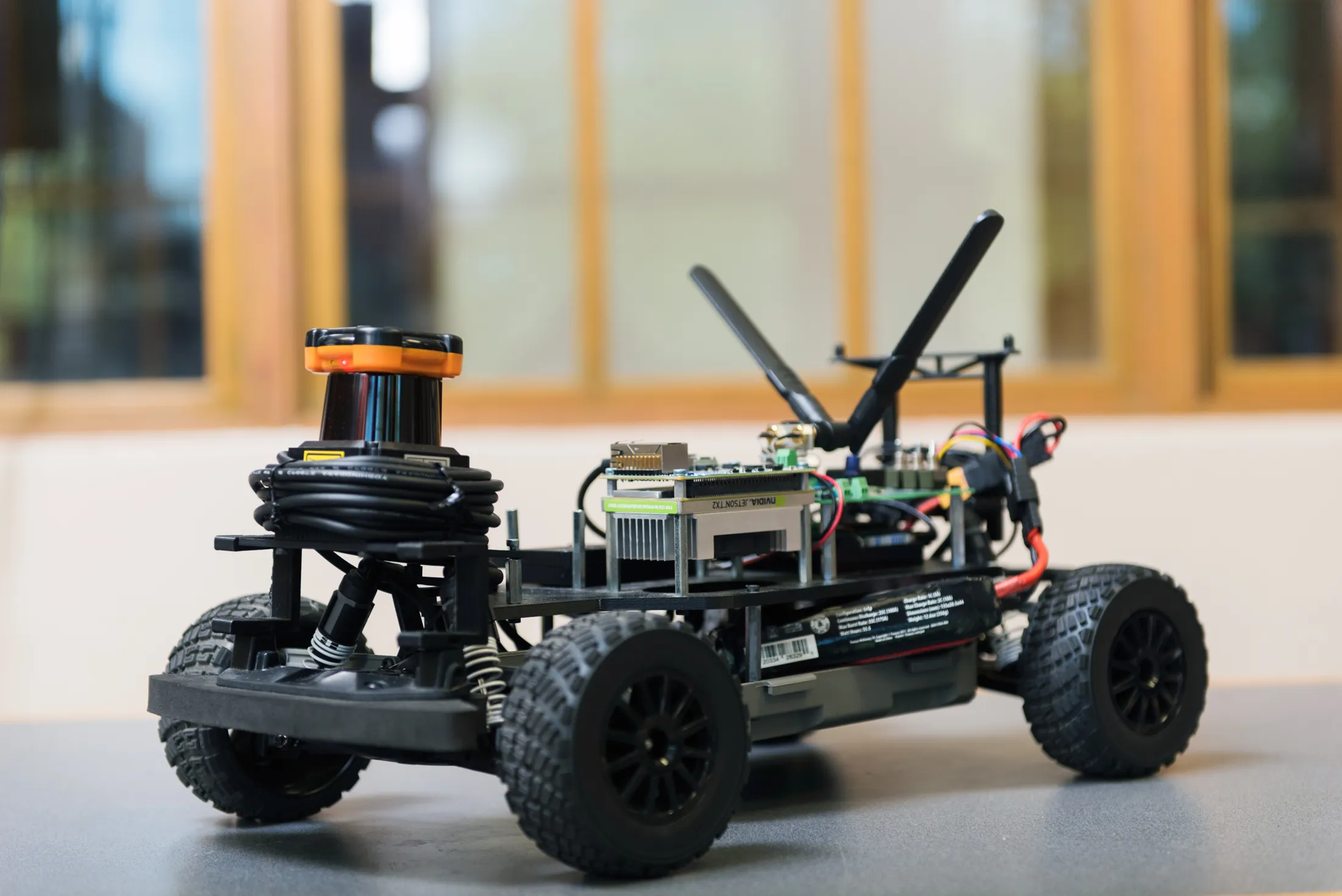}
	\caption{The vehicle used for the experiment (F1TENTH). Source: Steven Gong, \href{https://stevengong.co/notes/F1TENTH/}{https://stevengong.co/notes/F1TENTH/}.}
	\label{fig:realcar}
\end{figure}

\subsection{Experiment Results}
Following the promising simulation results, the RL agent was tested in real-world racing scenarios. F1TENTH car kit (Figure \ref{fig:realcar}) was chosen as the experiment vehicle. F1TENTH autonomous car is an open-source, 1/10th scale vehicle used for autonomous driving research and education. It's built on a radio-controlled (RC) car with components such as a sensor suite and power systems. It uses ROS (Robot Operating System) for the communication and is integrated with an Nvidia Jetson Xavier high-performance computer. 

The map used for the experiment is shown in Figure \ref{fig:realmap}. It consists of several sharp and smooth turns, as well as randomly placed obstacles along the way. Although the agent has been extensively trained on a variety of maps with differing curvatures, the test map presents a relatively unencountered features and challenges.

Surprisingly, the agent trained in the simulation has seamlessly adopted to the experimental tests without any further need of adjustments. The agent interacts with the map in a very similar way to how it does in the simulation. It takes the turns and safely avoids the obstacles in a very similar manner. It successfully centralizes itself in the track and slows down or stops when the area is ambiguous. 

The experimental track also includes features that the agent has not been trained on. This includes changing width of the racing track and parts that do not give away which turn to make, just based on the lidar readings (the issue of partial observability). At those points (marked on the map), the agent struggles to make the right decision and usually crashes to the borders.

Although the agent can mostly navigate safely through the track, because of the difference in the noise distribution between the simulation and the experiment, the steering stability is reduced. The agent sometimes tends to slightly oscillate between negative and positive steering, which is a behavior that is not observed in the simulation.

\printbibliography 
\end{document}